\documentclass{article}
\usepackage{ijcai25}

\usepackage{times}
\usepackage{soul}
\usepackage{url}
\usepackage[hidelinks]{hyperref}
\usepackage[utf8]{inputenc}
\usepackage[small]{caption}
\usepackage{graphicx}
\usepackage{amsmath}
\usepackage{amsthm}
\usepackage{booktabs}
\usepackage[switch]{lineno}
\usepackage{enumitem}
\usepackage{amsfonts}
\usepackage{algorithm}
\usepackage[noend]{algorithmic}
\usepackage[algo2e,noend]{algorithm2e}

\newcommand{\ours}{LIFTED}

\title{\ours: Multimodal Mixture-of-Experts for Clinical Trial Outcome Prediction}

\author{
Wenhao Zheng\footnote{Equal contribution.}$^1$\and
Liaoyaqi Wang\footnotemark[1]$^2$\and
Dongshen Peng$^1$\and
Hongxia Xu$^3$\and
Yun Li$^1$\and
Hongtu Zhu$^1$\and
Tianfan Fu$^4$\And
Huaxiu Yao\footnote{Corresponding author.}$^1$\\
\affiliations
$^1$UNC-Chapel Hill,
$^2$Johns Hopkins University,
$^3$Zhejiang University,
$^4$Rensselaer Polytechnic Institute\\
\emails
shenmishajing@gmail.com,
huaxiu@cs.unc.edu
}

\begin{document}

\maketitle

\begin{abstract}
    \vspace{-0.5em}
    The clinical trial is a pivotal and costly process, often spanning multiple years and requiring substantial financial resources. Therefore, the development of clinical trial outcome prediction models aims to exclude drugs likely to fail and holds the potential for significant cost savings. Recent data-driven attempts leverage deep learning methods to integrate multimodal data for predicting clinical trial outcomes. However, these approaches rely on manually designed modal-specific encoders, which limits both the extensibility to adapt new modalities and the ability to discern similar information patterns across different modalities. To address these issues, we propose a multimodal mixture-of-experts (\ours) approach for clinical trial outcome prediction. Specifically, \ours\ unifies different modality data by transforming them into natural language descriptions. Then, \ours\ constructs unified noise-resilient encoders to extract information from modal-specific language descriptions. Subsequently, a sparse Mixture-of-Experts framework is employed to further refine the representations, enabling \ours\ to identify similar information patterns across different modalities and extract more consistent representations from those patterns using the same expert model. Finally, a mixture-of-experts module is further employed to dynamically integrate different modality representations for prediction, which gives \ours\ the ability to automatically weigh different modalities and pay more attention to critical information. The experiments demonstrate that \ours\ significantly enhances performance in predicting clinical trial outcomes across all three phases compared to the best baseline, showcasing the effectiveness of our proposed key components.
\end{abstract}

\section{Introduction}
\label{sec:Introduction}

\begin{figure*}[h]
    \centering
    \includegraphics[width=0.92\textwidth]{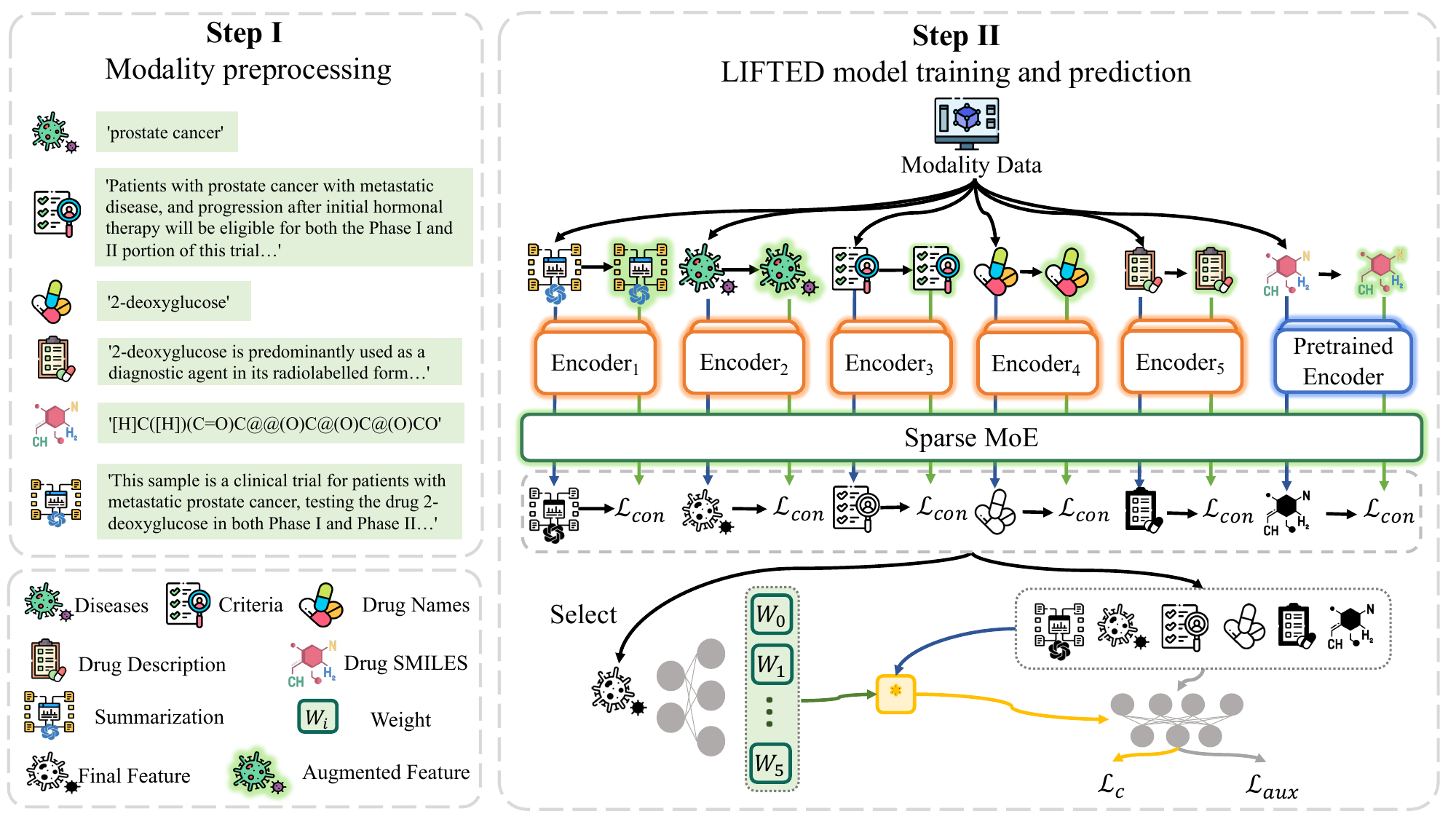}
    \caption{
        An overview of \ours. \textbf{Step 1}: Transforming multimodal data into natural language descriptions, where all modalities are converted into natural language descriptions to facilitate the representation extraction process of the transformer encoders. \textbf{Step 2}: Extract and combine representations from different modalities, where representations are extracted by the noise-resilient unified encoders and integrated by a Mixture-of-Experts (MoE) framework to make the final predictions.}
    \label{fig:model}
\end{figure*}

The clinical trial is a crucial step in the development of new treatments to demonstrate the safety and efficacy of the drug. Drugs must pass three trial phases involving human participants with target diseases before approval for manufacturing. However, the clinical trial is time-consuming and experiments expensive, taking multiple years and costing up to hundreds of millions of dollars~\cite{martin2017how}. In addition, the success rate of clinical trials is exceedingly low and many drugs fail to pass these clinical trials~\cite{wu2022cosbin,Huang20205545}. Therefore, the ability to predict clinical trial outcomes beforehand, allowing the exclusion of drugs with a high likelihood of failure, holds the potential to yield significant cost savings. Given the increasing accumulation of clinical trial data over the past decade (e.g., drug descriptions, and patient criteria), we can now leverage this wealth of data for the prediction of clinical trial outcomes.

Early attempts aim to improve the clinical trial outcome prediction results by modeling the components of the drugs (e.g., drug toxicity~\cite{gayvert2016data}, modeled the pharmacokinetics~\cite{qi2019predicting}). Recently, deep learning methods have been proposed for trial outcome predictions. For instance, Lo \textit{et al.}~\shortcite{lo2019learning} predicted drug approvals for 15 different disease groups by incorporating drug and clinical trial features into machine learning models. Fu \textit{et al.}~\shortcite{fu2022hint} proposed an interaction network leveraging multimodal data (e.g., molecule information, trial documents) to capture correlations for trial outcome predictions. However, this approach relies on modal-specific encoders to extract representations from different modal data, which require manually designed encoder structures and limit their extensibility when new modal data becomes available for use.

To address these issues, we aim to design a unified encoder to extract representations from various modalities, but it poses the following three challenges:
\begin{itemize}[leftmargin=*]
    \item \textbf{How to extract representations from different modalities with a unified encoder?} Different modalities are represented in various data formats. For instance, molecule information is typically depicted as graphs, while disease names rely on relationships between diseases. Therefore, a unified encoder structure should be capable of unifying these different formats to effectively extract information.
    \item \textbf{How to effectively utilize both the modality-independent information patterns and the modality-specific patterns to enhance the extracted representations?} Information across different modalities can be presented in both similar and different forms. For example, descriptions of a disease and corresponding drugs may mention the same symptoms, which can be extracted similarly. However, molecules and drug names represent information differently and should be extracted using distinct methods. Therefore, a method to dynamically identify similar information patterns across different modalities and direct them to the same encoder is also required.
    \item \textbf{How to integrate extracted information from different modalities?} Extracted representations from various modalities need to be integrated for predictions. However, the contribution of extracted information from different modalities may vary significantly between samples. For instance, in one patient, a specific disease, such as type 2 diabetes mellitus, which is difficult to treat, may strongly influence the final outcomes~\cite{wu2022cot}. In contrast, another patient's trial result may be primarily determined by the drugs they are prescribed, particularly if those medications have a high success rate in treating the disease. Hence, an approach to automatically weighting representations from different modalities is crucial.
\end{itemize}

To address those challenges, we propose an approach called mu\textbf{L}ti-modal m\textbf{I}x-of-experts \textbf{F}or ou\textbf{T}come pr\textbf{ED}iction (\textbf{LIFTED}), which extracts information from different modalities with a transformer based unified encoder, enhances the extracted features by a Sparse Mixture-of-Experts (SMoE) framework and integrates multimodal information with Mixture-of-Experts (MoE). Specifically, \ours\ unifies diverse multimodal features, even those in different formats, by converting them into natural language descriptions. Subsequently, we build a unified transformer-based encoder to extract representations from these modal-specific language descriptions and refine the representations with an SMoE framework. Here, the representations from different modalities are dynamically routed by a noisy top-k gating network to a portion of shared expert models, facilitating the extraction of similar information patterns. In addition, we introduce representation augmentation to enhance the resilience of transform-based encoders and the SMoE framework to potential data noise introduced during the data collection process. Furthermore, \ours\ treats the extracted representations from various modalities as distinct experts and utilizes a Mixture-of-Experts module to dynamically combine these multimodal representations for each example. This dynamic combination allows for the automatic assignment of higher weights to more crucial modalities. Finally, we evaluate \ours\ on the HINT benchmark~\cite{fu2022hint} and the CTOD benchmark~\cite{gao2024automaticallylabeling200blifesaving} to demonstrate the effectiveness of \ours\ and the effectiveness of our proposed components.

\section{Multimodal Mixture-of-Experts for Clinical Trial Outcome Prediction}

\label{sec:Method}

This section presents our proposed mu\textbf{L}ti-modal m\textbf{I}x-of-experts \textbf{F}or ou\textbf{T}come pr\textbf{ED}iction (\textbf{LIFTED}) method. The goal of \ours\ is to unify multimodal data using natural language descriptions and integrate this information within a Mixture-of-Experts (MoE) framework, as illustrated in Figure~\ref{fig:model}. To elaborate, we start by extracting specific modalities from the clinical trial dataset, subsequently transforming this multimodal data into natural language descriptions using a Large Language Model (LLM). Following this, we augment the embeddings of the language descriptions derived from these different modalities. We then feed both the original and augmented embeddings into transformer-based encoders for representation learning. Subsequently, an SMoE framework is utilized to route the embeddings from different modalities to different sets of experts, where similar information patterns in different modalities will be routed to the same experts while the different patterns will be routed to experts with more specialized knowledge. To enhance the robustness of encoders, we introduce a consistency loss that aligns the original representations with the augmented ones. Moving forward, we implement an MoE framework to integrate these representations for each trial, which originate from various modalities. Finally, these integrated representations are input into a classifier for prediction. Simultaneously, we introduce an auxiliary unimodal prediction loss to improve the quality of modal-specific representations.

\subsection{Transforming Multimodal Data into Natural Language Descriptions}
\label{subsec:preprocessing_of_the_tabular_data}

To build a unified encoder, the key challenge is how to unify multimodal data, which often have different structures for different modalities. For instance, molecule information is typically depicted as a graph, while disease names rely on relationships between different diseases~\cite{wu2022cot}. In \ours, we unify these different modality data by converting them into natural language descriptions. Specifically, we first format the input features into a key-value pair. After that, we use a prompt coupled with the corresponding key-value pair to ask an LLM to generate a natural language description for our input. Subsequently, these descriptions will be fed into a unified tokenizer for further encoding, except the SMILES string modality, which is tokenized by a specifically designed tokenizer to enhance the representation of molecule information. The first two steps, linearization and prompting, are detailed below:

\paragraph{Linearization.} In linearization, we format each data point $x_{i,k}$ of trial $i$ and modality $k$ into a key-value pair. In this pair, the key of each element represents the feature name $c_{i,k}$, and the corresponding value is $x_{i,k}$. This can be formulated as follows:
\begin{equation}
    \label{eq:linearization_single_modality}
    \mathrm{Linearize}(x_{i,k})=\{c_{i,k}: x_{i,k}\}.
\end{equation}

\begin{figure}
    \centering
    \includegraphics[width=\columnwidth]{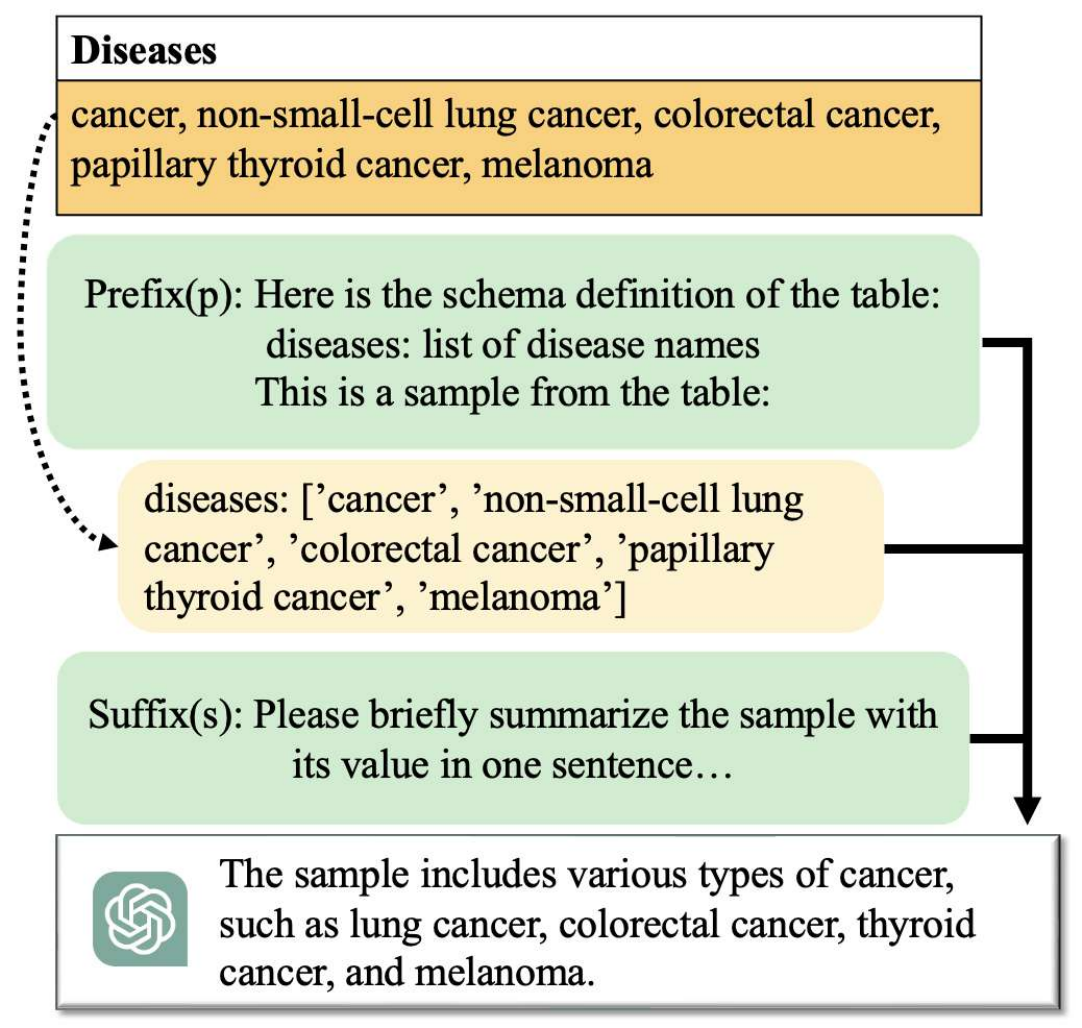}
    \caption{Processes of the linearization and the prompting.}
    \label{fig:prompting}
\end{figure}

\paragraph{Prompting.} As depicted in Figure~\ref{fig:prompting}, the prompts we use to communicate with the LLM consist of three components: a prefix $p$ to describe the schema of the input features, the linearization and a suffix $s$ to instruct the LLM on how to describe the input data point in natural language. Given the prompts, the LLM will generate a readable and concise natural description $z_{i,k}$, which can be formulated as:
\begin{equation}
    \label{eq:prompting}
    z_{i,k}=\mathrm{LLM}(p, \mathrm{Linearize}(x_{i,k}), s).
\end{equation}
For instance, given the linearization of disease modality, ``diseases: ['cancer', 'non-small-cell lung cancer', 'colorectal cancer', 'papillary thyroid cancer', 'melanoma']'', the LLM will generate the natural language description like ``The sample includes various types of cancer, such as lung cancer, colorectal cancer, thyroid cancer, and melanoma.''

In addition to transforming existing modality data to natural language descriptions, we also generate a new summarization modality to provide an overall description of the whole trial in a similar way. The only difference for generating the summarization is that we concatenate the linearization of all modalities as input to provide the information of the whole trial as (see the prompt in Appendix A):
\begin{equation}
    \label{eq:linearization}
    \begin{aligned}
        z_{i,0}=        \mathrm{LLM}(p,           & \;\mathrm{Linearize}(x_i), s),   \\
        \mathrm{where}\;\;\mathrm{Linearize}(x_i) & =  \{c_{i,k}: x_{i,k}\}^K_{k=1}.
    \end{aligned}
\end{equation}
By this, all the information is summarized in text format.

\subsection{Representation Learning and Refinement}
\label{subsec:the_structure_of_the_transformer_encoders}

After transforming multimodal data into natural language descriptions, we build $K+1$ transformer-based encoders on top of these descriptions. Specifically, each modality description $z_{i,k}$ is tokenized into a sequence of tokens $\{z_{i,k}^t\}_{t=1}^T$ with length $T$ by a tokenizer $\mathcal{T}$ and embedded into a sequence of embeddings $\{u_{i,k}^t\}_{t=1}^T$ by a modal-specific embedding layer $\mathcal{E}_k$ first, and then they are added by the position embeddings $\mathrm{pos}^t$ and fed into the correponding modal-specific transformer encoder $\mathcal{F}_k$ coupled with a learnable token $[cls]_k$ to get encoded representation $U_{i,k}$. The encoding process can be formulated as follows:

\begin{equation}
    \label{eq:encoder}
    \begin{aligned}
         & \{z_{i,k}^t\}_{t=1}^T =  \mathcal{T}(z_{i,k})                        \\
         & u_{i,k}^t =             \mathcal{E}_k(z_{i,k}^t)                     \\
         & U_{i,k}   =            \mathcal{F}_k(\{u_{i,k}^t + pos^t\}_{t=0}^T),
    \end{aligned}
\end{equation}
where, we define $u_{i,k}^0 \equiv [cls]_k$.

Furthermore, to equip \ours\ with the capability to dynamically identify similar information patterns across different modalities and route them to the same encoder, we employ a Sparse Mixture-of-Experts (SMoE) framework to further refine the extracted representations. The encoded representations $U_{i,k}$ from different modalities will be dynamically routed by a modality-independent noisy top-k gating network $\mathcal{G}$ to a subset of shared expert models $\{\mathcal{R}^r\}_{r=1}^R$ to facilitate the extraction of similar information patterns, following the original design of SMoE~\cite{shazeer2017outrageously}. The whole process can be formulated as follows:
\begin{equation}
    \label{eq:smoe_gating_network}
    \begin{aligned}
        \mathcal{G}(U_{i,k}) =  & \mathrm{Softmax}(\mathrm{TopK}(\mathcal{P}(U_{i,k}), k))                    \\
        \mathcal{P}(U_{i,k}) =  & U_{i,k} \cdot W_g + \mu \mathrm{Softplus}(U_{i,k} \cdot W_{\mathrm{noise}}) \\
        \mathrm{TopK}(v, k)_j = & \left\{ \begin{aligned}
                                               & v_j,     & \text{if $v_j$ in the top $k$ elements of $v$} \\
                                               & -\infty, & \text{otherwise}
                                          \end{aligned} \right.
    \end{aligned}
\end{equation}
\noindent where the $\mu$ is random noise sampled from a standard normal distribution, $W_g$ is a learnable weight matrix shared through different modalities and $W_{\mathrm{noise}}$ is another learnable noise matrix to control the amount of noise per component. Subsequently, the encoded representations $U_{i,k}$ will be routed only to the shared expert models $\{\mathcal{R}^r\}_{r=1}^R$ with top-k gating scores generated by the gating network $\mathcal{G}$. The refined representations $\tilde{U}_{i,k}$ can then be calculated by combining the encoding results from the top-k expert models with their corresponding gating scores. The whole process can be formulated as follows:

\begin{equation}
    \label{eq:smoe}
    \begin{aligned}
        \tilde{U}_{i,k} = & \mathcal{G}(U_{i,k}) \cdot \mathcal{R}(U_{i,k}) = & \sum_{r=1}^{R}\mathcal{G}^r(U_{i,k})\mathcal{R}^r(U_{i,k}).
    \end{aligned}
\end{equation}

\subsection{Consistent Representation Augmentation}
\label{subsec:data_augmentation_and_consistency_loss}

However, building informative modal-specific encoders and the SMoE framework solely from these modal-specific natural language descriptions remains challenging, primarily due to potential data noise introduced during the data collection process. To make the encoders and the SMoE framework more robust to the noise in the data, we augment the embeddings $u_{i,k}^t$ with a minor perturbation to $v_{i,k}^t$ and add a consistency loss to require the encoders and the SMoE framework insensitive to small perturbation.

\paragraph{Representation Augmentation.} To perform representation augmentation, we begin by considering each embedding vector $u^t_{i,k} \in \mathbb{R}^L$, where $L$ represents the number of elements $\{m_{l}\}_{l=1}^L$. We randomly select a subset of these elements from $u^t_{i,k}$ with a probability $p$ for perturbation, while leaving the remaining elements unchanged. For simplicity, we will omit the subscript and superscript for $m_l$ specific to the embedding $u^t_{i,k}$. Next, we proceed to sample a small value $\alpha_l$ from a uniform distribution $\mathrm{Uniform}(-\lambda, \lambda)$ for each selected element. Here, $\lambda$ serves as a hyperparameter that controls the magnitude of the minor perturbation. Following this, each selected element is multiplied by $\exp(\alpha_l)$ to apply the perturbation. This process can be expressed as:
\begin{equation}
    \label{eq:augmentation}
    \hat{m}_l=\left\{
    \begin{aligned}
         & \exp(\alpha_l) * m_l, & m_l\ \text{is selected} \\
         & m_l,                  & \text{otherwise}        \\
    \end{aligned}
    \right.
\end{equation}
After his, we get the perturbed vector $v^t_{i,k}=\{\hat{m}_{l}\}_{l=1}^L$.

\paragraph{Consistency Loss.} After the representation augmentation step, we obtain a sequence of perturbed embeddings $\{v_{i,k}^t\}_{t=1}^T$ derived from the original embeddings $\{u_{i,k}^t\}_{t=1}^T$. These perturbed embeddings are then input into the encoder $\mathcal{F}_k$ and the SMoE framework to generate the encoded representation $\tilde{V}_{i,k}$. In order to ensure the robustness of the encoded embeddings, we introduce a consistency loss $\mathcal{L}_{con}$ to control the disparity between the encoded representation of the original embeddings and the augmented embeddings. This consistency loss can be formulated as follows
\begin{equation}
    \label{eq:consistency_loss}
    \begin{aligned}
        \mathcal{L}_{con} =  \frac{1}{N(K+1)}\sum_{k=0}^{K} & \sum_{i=1}^{N} \left\| \tilde{U}_{i,k} - \tilde{V}_{i,k} \right\|^2_F, \\
        \mathrm{where} \;\;  \tilde{V}_{i,k} =              & \mathcal{G}(V_{i,k}) \cdot \mathcal{R}(V_{i,k}),                       \\
        V_{i,k} =                                           & \mathcal{F}_k(\{v_{i,k}^t + pos^t\}_{t=0}^T),
    \end{aligned}
\end{equation}
where we define $v_{i,k}^0 \equiv [cls]_k$.

\subsection{Integrating Multimodal Information}
\label{subsec:intergrating_multimodal_information}
After obtaining representations from different modalities, the model's ability to integrate these multimodal representations and discern the patient-specific importance of each modality becomes crucial for precise predictions. As illustrated in Figure~\ref{fig:model}, we employ a Mixture-of-Experts (MoE) framework to dynamically integrate multimodal representations. In this framework, we treat the extracted representations from various modalities as distinct experts.

Concretely, for each example $i$, we start by concatenating the extracted representations from the selected modalities and then feed them into a fully connected layer denoted as $\mathcal{C}$ to calculate the modality importance weights $W_{i,k}$ for each modality. In our implementation, we exclusively utilize the disease modality to generate these importance weights, as knowing the patient's disease allows us to determine which modality should receive emphasis. Subsequently, we multiply these weights by their corresponding representations $\{U_{i,k}\}_{k=0}^{K}$ and aggregate them to obtain the integrated representation $U_i$. The process can be formulated as follows:
\begin{equation}
    \label{eq:moe_method}
    \begin{aligned}
        W_{i,k}= & \mathrm{Softmax}(\mathcal{C}(\oplus_{j \in \mathcal{J}} U_{i,j}) * \gamma_k) \\
        U_i =    & \sum_{k=0}^K W_{i,k}*\tilde{U}_{i,k},
    \end{aligned}
\end{equation}
where the $\oplus$ is the concatenate operation along the representation dimension and the $\mathcal{J}$ is the set of selected modalities. $\gamma_k$ is a learnable modal-specific temperature factor.

Following this, we make the prediction $\hat{y}_i$ by inputting the integrated representation $U_i$ into the classifier $\mathcal{H}$. The classification loss $\mathcal{L}_c$ is defined as follows:
\begin{equation}
    \label{eq:classification_loss}
    \begin{aligned}
        \mathcal{L}_c =       \frac{1}{N}  \sum_{i=1}^N \ell(\hat{y}_i, y_i), \ \ \mathrm{where}\ \  \hat{y}_i=  \mathcal{H}(U_i)
    \end{aligned}
\end{equation}
where the $y_i$ is the ground truth label for sample $i$ and the loss term $\ell$ is the cross entropy loss.

To ensure that the unimodal representations are of high quality and consistently contribute to the final prediction, we introduce an auxiliary loss to align the representations from different modalities. Similar to the classification loss $\mathcal{L}_c$, the auxiliary loss $\mathcal{L}_{aux}$ is calculated as the sum of uni-modal prediction losses, which can be formulated as follows:

\begin{equation}
    \label{eq:auxiliary_loss}
    \begin{aligned}
                            & \hat{y}_{i,k}=  \mathcal{H}(U_{i,k})                                  \\
        \mathcal{L}_{aux} = & \frac{1}{N(K+1)} \sum_{k=0}^{K}\sum_{i=1}^N \ell(\hat{y}_{i,k}, y_i).
    \end{aligned}
\end{equation}
Finally, the overall loss $\mathcal{L}$ is defined as:
\begin{equation}
    \label{eq:overall_loss}
    \mathcal{L}=\mathcal{L}_c+\eta_1 \mathcal{L}_{con}+\eta_2 \mathcal{L}_{aux},
\end{equation}
where $\eta_1$ and $\eta_2$ are hyperparameters to balance these loss terms. The whole algorithm is illustrated in Alg.~\ref{alg:model}.

\begin{algorithm}[h]
    \caption{Training Pipeline of \ours}
    \label{alg:model}
    \SetKwData{Left}{left}\SetKwData{This}{this}\SetKwData{Up}{up} \SetKwFunction{Union}{Union}\SetKwFunction{FindCompress}{FindCompress} \SetKwInOut{Input}{Input}\SetKwInOut{Output}{Output}

    \Input{Training dataset $\mathcal{D}=\{x_i, y_i\}_{i=1}^N$}

    \emph{step 1. Transforming Multimodal Data into Natural Language Descriptions}\;
    \For{$i\leftarrow 1$ \KwTo $N$}{
        \For{$k\leftarrow 1$ \KwTo $K$}{
            $x_{i,k} \leftarrow \mathrm{LLM}(p, \mathrm{Linearize}(x_{i,k}), s)$}
        $x_{i,0} \leftarrow \mathrm{LLM}(p, \mathrm{Linearize}(x_i), s)$
    }
    \emph{step 2. Train \ours}\;
    \ForEach{minibatch $\mathcal{B}$ in dataset $\mathcal{D}$}{
    \For{$i\leftarrow 1$ \KwTo batch size $b$}{
    \For{$k\leftarrow 0$ \KwTo $K$}{
    $u_{i,k}^t \leftarrow \mathcal{E}_k(\mathcal{T}(z_{i,k}))$ \\
    $v_{i,k}^t \leftarrow \hat{u}_{i,k}^t$ as Equation~\ref{eq:augmentation} \\
    $U_{i,k} \leftarrow \mathcal{F}_k(\{u_{i,k}^t + pos^t\}_{t=0}^T)$, \\
    $\tilde{U}_{i,k} \leftarrow \mathcal{G}(U_{i,k}) \cdot \mathcal{R}(U_{i,k})$, \\
    $V_{i,k} \leftarrow \mathcal{F}_k(\{v_{i,k}^t + pos^t\}_{t=0}^T)$, \\
    $\tilde{V}_{i,k} \leftarrow \mathcal{G}(V_{i,k}) \cdot \mathcal{R}(V_{i,k})$}
    Fuse representations with MoE method as~\eqref{eq:moe_method}
    }
    Compute the losses $\mathcal{L}_{con}$, $\mathcal{L}_c$, $\mathcal{L}_{aux}$ and $\mathcal{L}$\\
    Optimize the parameters $\theta$ of \ours\
    }
\end{algorithm}

\section{Experiments}
\label{sec:Experiment}

In this section, we evaluate the performance of \ours\, aiming to answer the following questions: \textbf{Q1}: Compared to the existing methods with modal-specific encoders, can \ours\ achieve better performance with the unified transformer encoders? \textbf{Q2}: Do the key components, including the multimodal data integration component and the representation augmentation component, of \ours\ boost the performance? \textbf{Q3}: Does the Sparse Mixture-of-Experts framework route similar information patterns in different modalities to the same expert models correctly? \textbf{Q4}: Does the Mixture-of-Experts approach precisely measure the importance of different modalities for each patient?

\begin{table*}[h]
    \caption{The clinical trial outcome performance (\%) of \ours\ and baselines on HINT dataset and CTOD dataset. The results are averaged on 30 independent runs with different random seeds. $^\dagger$: The results of the HINT and SPOT methods were obtained by running their released codes. The best results and second best results are \textbf{bold} and \underline{underlined}, respectively. We observe that \ours\ consistently outperforms all other methods over all three phases.}
    \label{tab:performance_phase}
    \resizebox{\linewidth}{!}{
        \begin{tabular}{l|c|c|c|c|c|c|c|c|c|c|c|c|c|c|c|c|c|c}
            \toprule
                            & \multicolumn{9}{c|}{\textbf{HINT}}           & \multicolumn{9}{c}{\textbf{CTOD}}                                                                                                                                                                                                                                                                                                                                                                                                                                                                            \\
            \midrule
                            & \multicolumn{3}{c|}{\textbf{Phase I Trials}} & \multicolumn{3}{c|}{\textbf{Phase II Trials}} & \multicolumn{3}{c|}{\textbf{Phase III Trials}} & \multicolumn{3}{c|}{\textbf{Phase I Trials}} & \multicolumn{3}{c|}{\textbf{Phase II Trials}} & \multicolumn{3}{c}{\textbf{Phase III Trials}}                                                                                                                                                                                                                                                                \\
            \cmidrule{2-19}
            \textbf{Method} & \textbf{PR}                                  & \textbf{F1}                                   & \textbf{ROC}                                   & \textbf{PR}                                  & \textbf{F1}                                   & \textbf{ROC}                                  & \textbf{PR}        & \textbf{F1}        & \textbf{ROC}       & \textbf{PR}        & \textbf{F1}        & \textbf{ROC}       & \textbf{PR}        & \textbf{F1}         & \textbf{ROC}       & \textbf{PR}        & \textbf{F1}          & \textbf{ROC}       \\
            \midrule
            LR              & $50.0$                                       & $60.4$                                        & $52.0$                                         & $56.5$                                       & $55.5$                                        & $58.7$                                        & $68.7$             & $69.8$             & $65.0$             & $55.6$             & $58.5$             & $51.5$             & $56.0$             & $63.6$              & $56.7$             & $75.0$             & $73.8$               & $51.9$             \\
            RF              & $51.8$                                       & $62.1$                                        & $52.5$                                         & $57.8$                                       & $56.3$                                        & $58.8$                                        & $69.2$             & $68.6$             & $66.3$             & $58.3$             & $70.0$             & $54.0$             & $60.8$             & $70.9$              & $56.0$             & $75.1$             & $85.1$               & $50.7$             \\
            XGBoost         & $51.3$                                       & $62.1$                                        & $51.8$                                         & $58.6$                                       & $57.0$                                        & $60.0$                                        & $69.7$             & $69.6$             & $66.7$             & $57.4$             & $68.0$             & $50.5$             & $61.2$             & $70.6$              & $61.3$             & $78.7$             & $84.9$               & $57.1$             \\
            AdaBoost        & $51.9$                                       & $62.2$                                        & $52.6$                                         & $58.6$                                       & $58.3$                                        & $60.3$                                        & $70.1$             & $69.5$             & $67.0$             & $58.8$             & $66.7$             & $53.5$             & $60.2$             & $68.9$              & $55.2$             & $79.5$             & $84.6$               & $56.9$             \\
            kNN+RF          & $53.1$                                       & $62.5$                                        & $53.8$                                         & $59.4$                                       & $59.0$                                        & $59.7$                                        & $70.7$             & $69.8$             & $67.8$             & $58.4$             & $70.3$             & $54.7$             & $61.2$             & $70.9$              & $56.6$             & $76.8$             & $85.2 $              & $52.5 $            \\
            \midrule
            FFNN            & $54.7$                                       & $63.4$                                        & $55.0$                                         & $60.4$                                       & $59.9$                                        & $61.1$                                        & $74.7$             & $74.8$             & $68.1$             & $55.2$             & $58.4$             & $48.7$             & $57.2$             & $66.0$              & $53.0$             & $76.1$             & $79.6$               & $52.6$             \\
            MMF (early)     & $60.6$                                       & $59.4$                                        & $54.4$                                         & $60.2$                                       & $62.6$                                        & $60.7$                                        & $85.5$             & $81.5$             & $70.6$             & $61.2$             & $64.2$             & $56.2$             & $64.0$             & $70.1$              & $59.0$             & $83.3$             & $84.4$               & $64.6$             \\
            MMF (late)      & $63.4$                                       & $67.5$                                        & $59.0$                                         & \underline{$62.9$}                           & $63.0$                                        & $62.6$                                        & \underline{$86.9$} & \underline{$83.1$} & \underline{$71.8$} & $63.0$             & $70.5$             & $57.4$             & \underline{$65.7$} & \underline{$71.4$ } & $60.3$             & \underline{$83.5$} & $85.2$               & $65.8$             \\
            HINT$^\dagger$  & $58.4$                                       & $68.2$                                        & $62.1$                                         & $59.1$                                       & $63.9$                                        & $62.8$                                        & \underline{$85.9$} & $80.9$             & $70.8$             & $63.4$             & \underline{$71.0$} & $57.6$             & $64.6$             & $71.2$              & $60.6$             & $81.8$             & \underline{$85.7$  } & $60.8$             \\
            SPOT$^\dagger$  & \underline{$69.8$}                           & \underline{$68.4$}                            & \underline{$64.6$}                             & $62.6$                                       & \underline{$64.3$}                            & \underline{$63.0$}                            & $81.7$             & $81.0$             & $71.0$             & \underline{$66.8$} & $70.0$             & \underline{$62.3$} & $64.5$             & $71.8$              & \underline{$58.8$} & $83.2$             & $77.6$               & \underline{$67.5$} \\
            \midrule
            \textbf{\ours}  & $\textbf{70.7}$                              & $\textbf{71.6}$                               & $\textbf{64.9}$                                & $\textbf{69.8}$                              & $\textbf{66.2}$                               & $\textbf{65.1}$                               & $\textbf{88.3}$    & $\textbf{83.8}$    & $\textbf{73.5}$    & $\textbf{69.7}$    & $\textbf{71.8}$    & $\textbf{63.4}$    & $\textbf{67.7}$    & $\textbf{72.0}$     & $\textbf{63.0}$    & $\textbf{86.7}$    & $\textbf{85.9}$      & $\textbf{69.5}$    \\
            \bottomrule
        \end{tabular}}
\end{table*}

\subsection{Experimental Setup}
\label{subsec:Experiment_Setup}

\noindent \textbf{Dataset Descriptions.} We evaluate our method and other baselines on the HINT dataset~\cite{fu2022hint,chen2024uncertainty} and CTOD dataset~\cite{gao2024automaticallylabeling200blifesaving}, covering Phases I, II and III trials.
More details of the HINT dataset and the CTOD dataset are shown in Appendix C.

\paragraph{Baselines.} We compare \ours\ with both machine learning methods and deep learning models, such as Feedforward Neural Network (FFNN)~\cite{shen2023genocraft}, Multi Modal Fusion (MMF), HINT~\cite{fu2022hint,wang2024twin}, SPOT~\cite{wang2023spot}.
More details of those baselines are presented in Appendix B.

\paragraph{Evaluation Metrics.} Following Fu \textit{et al.}~\shortcite{fu2022hint} and Chen \textit{et al.}~\shortcite{chen2024uncertainty}, we use F1 score, PR-AUC, and ROC-AUC to measure the performance of all methods. For all these metrics, higher scores indicate better performance.

\begin{table}[h]
    \vspace{-1em}
    \caption{The clinical trial outcome prediction performance (\%) of \ours\ and variants without certain key component. The best results are \textbf{bold}. \ours\ outperforms all variants, showcasing the effectiveness of our proposed components.}
    \vspace{-1em}
    \label{tab:data_augment_ablation_study}
    \resizebox{\linewidth}{!}{
        \begin{tabular}{l|c|c|c|c|c|c|c|c|c}
            \toprule
            \multicolumn{10}{c}{\textbf{HINT}}                                                                                                                                                                                                                                       \\
            \midrule
                            & \multicolumn{3}{c|}{\textbf{Phase I Trials}} & \multicolumn{3}{c|}{\textbf{Phase II Trials}} & \multicolumn{3}{c}{\textbf{Phase III Trials}}                                                                                                           \\
            \cmidrule{2-10}
            \textbf{Method} & \textbf{PR}                                  & \textbf{F1}                                   & \textbf{ROC}                                  & \textbf{PR}     & \textbf{F1}     & \textbf{ROC}    & \textbf{PR}     & \textbf{F1}     & \textbf{ROC}  \\
            \midrule

            \ours-aug
                            & $68.4$                                       & $69.8$                                        & $64.8$
                            & $69.5$                                       & $66.0$                                        & $64.3$
                            & $86.9$                                       & $82.4$                                        & $72.1$                                                                                                                                                  \\

            \ours-aux       & $69.0$                                       & $71.2$                                        & $63.7$                                        &
            $69.6$          & $64.5$                                       & $64.6$                                        &
            $87.4$          & $82.8$                                       & $71.1$                                                                                                                                                                                                  \\

            \ours-LLM       & $68.5$                                       & $70.8$                                        & $64.0$                                        & $69.7$          & $64.9$          & $65.0$          & $86.7$          & $82.7$          & $70.8$        \\

            \ours-gating    & 69.9                                         & 71.3                                          & 64.9
                            & 69.7                                         & 65.5                                          & 65.0
                            & 87.0                                         & 82.7                                          & 72.4                                                                                                                                                    \\
            \midrule

            \textbf{\ours}  & $\textbf{70.7}$                              & $\textbf{71.6}$                               & $\textbf{64.9}$                               & $\textbf{69.8}$ & $\textbf{66.2}$ & $\textbf{65.1}$ & $\textbf{88.3}$ & $\textbf{83.8}$ & \textbf{73.5} \\
            \midrule\midrule
            \multicolumn{10}{c}{\textbf{CTOD}}                                                                                                                                                                                                                                       \\
            \midrule
                            & \multicolumn{3}{c|}{\textbf{Phase I Trials}} & \multicolumn{3}{c|}{\textbf{Phase II Trials}} & \multicolumn{3}{c}{\textbf{Phase III Trials}}                                                                                                           \\
            \cmidrule{2-10}
            \textbf{Method} & \textbf{PR}                                  & \textbf{F1}                                   & \textbf{ROC}                                  & \textbf{PR}     & \textbf{F1}     & \textbf{ROC}    & \textbf{PR}     & \textbf{F1}     & \textbf{ROC}  \\
            \midrule

            {\ours-aug}     &
            $65.1$          & $70.6$                                       & {$61.9$}                                      &
            {$67.1$}        & $70.5$                                       & {$62.2$}                                      &
            {$85.2$}        & $82.4$                                       & $67.8$                                                                                                                                                                                                  \\

            {\ours-aux}     &
            $64.4$          & {$71.1$}                                     & $60.0$                                        & $60.0$                                        & $71.0$          & $54.9$          & $83.8$          & $85.5$          & $64.6$                          \\

            {\ours-LLM}     &
            $65.1$          & {$71.4$}                                     & $58.5$                                        & $66.3$                                        & $71.1$          & $61.2$          & $83.9$          & $85.4$          & $65.3$                          \\

            {\ours-gating}  &
            {$67.1$}        & {$71.4$}                                     & $60.8$                                        &
            $65.1$          & $70.4$                                       & $61.9$                                        &
            $85.0$          & {$85.8$}                                     & {$68.8$}                                                                                                                                                                                                \\
            \midrule
            \textbf{\ours}  &
            $\textbf{67.7}$ & $\textbf{71.6}$                              & $\textbf{62.3}$                               &
            $\textbf{67.7}$ & $\textbf{72.0}$                              & $\textbf{63.0}$                               &
            $\textbf{86.7}$ & $\textbf{85.9}$                              & $\textbf{69.5}$                                                                                                                                                                                         \\

            \bottomrule
        \end{tabular}}
    \vspace{-1em}
\end{table}

\begin{table}[h]
    \vspace{-1em}
    \caption{Performance analysis of multimodal data integration. The best results and second best results are \textbf{bold} and \underline{underlined}, respectively.}
    \vspace{-1em}
    \label{tab:moe_method_ablation_study}
    \resizebox{\linewidth}{!}{
        \begin{tabular}{l|c|c|c|c|c|c|c|c|c}
            \toprule
            \multicolumn{10}{c}{\textbf{HINT}}                                                                                                                                                                                                                                                                \\
            \midrule
                                 & \multicolumn{3}{c|}{\textbf{Phase I Trials}} & \multicolumn{3}{c|}{\textbf{Phase II Trials}} & \multicolumn{3}{c}{\textbf{Phase III Trials}}                                                                                                                               \\
            \cmidrule{2-10}
                                 & \textbf{PR}                                  & \textbf{F1}                                   & \textbf{ROC}                                  & \textbf{PR}        & \textbf{F1}        & \textbf{ROC}       & \textbf{PR}        & \textbf{F1}        & \textbf{ROC}       \\
            \midrule
            Summarization        & $63.2$                                       & $69.9$                                        & $57.8$                                        & $66.1$             & $61.2$             & $61.2$             & $85.1$             & $80.7$             & $66.6$             \\
            Drugs                & $62.1$                                       & $67.2$                                        & $57.9$                                        & $60.5$             & $62.3$             & $55.8$             & $83.8$             & $81.7$             & $63.8$             \\
            Disease              & $65.3$                                       & $67.3$                                        & $59.7$                                        & \underline{$68.0$} & $59.9$             & $62.4$             & \underline{$86.0$} & $80.5$             & \underline{$69.1$} \\
            Description          & $55.5$                                       & \underline{$71.3$}                            & $50.2$                                        & $55.5$             & $ 0.0$             & $50.0$             & $74.9$             & $\textbf{85.7}$    & $49.7$             \\
            SMILES               & $62.8$                                       & $69.6$                                        & $58.5$                                        & $59.3$             & $58.3$             & $54.9$             & $76.1$             & $83.6$             & $51.1$             \\
            Criteria             & \underline{$68.0$}                           & $70.5$                                        & \underline{$63.1$}                            & $67.6$             & \underline{$64.4$} & \underline{$63.0$} & $83.7$             & $82.7$             & $65.0$             \\
            \midrule
            \textbf{All (\ours)} & $\textbf{70.7}$                              & $\textbf{71.6}$                               & $\textbf{64.9}$                               & $\textbf{69.8}$    & $\textbf{66.2}$    & $\textbf{65.1}$    & $\textbf{88.3}$    & \underline{$83.8$} & $\textbf{73.5}$    \\
            \midrule\midrule
            \multicolumn{10}{c}{\textbf{CTOD}}                                                                                                                                                                                                                                                                \\
            \midrule
                                 & \multicolumn{3}{c|}{\textbf{Phase I Trials}} & \multicolumn{3}{c|}{\textbf{Phase II Trials}} & \multicolumn{3}{c}{\textbf{Phase III Trials}}                                                                                                                               \\
            \cmidrule{2-10}
                                 & \textbf{PR}                                  & \textbf{F1}                                   & \textbf{ROC}                                  & \textbf{PR}        & \textbf{F1}        & \textbf{ROC}       & \textbf{PR}        & \textbf{F1}        & \textbf{ROC}       \\
            \midrule
            Summarization        &
            $61.6$               & $70.9$                                       & $56.9$                                        &
            $65.4$               & \underline{$71.5$}                           & $61.1$                                        &
            $84.1$               & {$84.8$}                                     & $64.9$                                                                                                                                                                                                                      \\
            Drugs                &
            $60.8$               & $70.6$                                       & $57.6$                                        &
            $58.0$               & $71.3$                                       & $53.8$                                        &
            $77.3$               & $85.2$                                       & $54.5$                                                                                                                                                                                                                      \\
            Disease              &
            \underline{$65.2$}   & $70.2$                                       & \underline{$61.3$}                            &
            \underline{$68.3$}   & $71.1$                                       & \underline{$62.6$}                            &
            \underline{$85.3$}   & $85.5$                                       & \underline{$68.4$}
            \\
            Description          &
            $56.6$               & $70.7$                                       & $52.3$                                        &
            $56.0$               & $71.3$                                       & $52.4$                                        &
            $76.5$               & \underline{$85.7$}                           & $52.9$                                                                                                                                                                                                                      \\
            SMILES               &
            $61.6$               & $70.6$                                       & $56.0$                                        &
            $58.0$               & \underline{$71.5$}                           & $53.1$                                        &
            $75.1$               & {$85.6$}                                     & $50.8$                                                                                                                                                                                                                      \\
            Criteria             &
            $60.5$               & $71.2$                                       & $55.8$                                        &
            $59.9$               & $71.2$                                       & $53.9$                                        &
            $75.7$               & {$84.6$}                                     & $52.5$                                                                                                                                                                                                                      \\
            \midrule
            \textbf{All (\ours)} &
            $\textbf{67.7}$      & $\textbf{71.6}$                              & $\textbf{62.3}$                               &
            $\textbf{67.7}$      & $\textbf{72.0}$                              & $\textbf{63.0}$                               &
            $\textbf{86.7}$      & $\textbf{85.9}$                              & $\textbf{69.5}$                                                                                                                                                                                                             \\
            \bottomrule
        \end{tabular}}
    \vspace{-1.5em}
\end{table}

\subsection{Overall Performance}
\label{subsec:Results}

We conduct experiments to evaluate the performance of \ours\ on all three phases trails, compared to our baselines. The trial outcome prediction results of all models are reported in Table~\ref{tab:performance_phase}.
We first observed that the deep learning-based methods and the methods designed for clinical trial outcome prediction outperform the machine learning based methods with a significant performance gap, especially on the HINT dataset, showcasing the powerful ability to extract critical information from different modalities in various formats of the deep learning encoders and those encoders specifically designed to extract representation hidden in the clinical trial records. This observation is not surprising, since the critical information of different modalities is represented in different ways, which is hard to extract for those traditional machine learning methods or those deep learning encoders that are not designed for clinical trial outcome prediction. Nevertheless, \ours\ consistently outperforms all other methods over all three phases, verifying its effectiveness in unifying different modalities and dynamically integrating them within the MoE.

\subsection{Ablation Study}
\label{subsubsec:auxiliary_loss}

In this section, we perform comprehensive ablation studies to demonstrate the effectiveness of our key components, including the representation augmentation, the auxiliary loss and the modalities used to generate weights in the Mixture-of-Experts (MoE) framework.
\begin{itemize}[leftmargin=*]
    \item \ours-aug: In \ours-aug, the representation augmentation component and the consistency loss are removed. Representations from different modalities are directly fed into the multimodal data integration component without the constraint of robustness to the noise in data.
    \item \ours-aux: In \ours-aux, we remove the auxiliary loss component. Representations from different modalities are no longer required to make consistent predictions with the final representation.
    \item \ours-LLM: In \ours-LLM, we remove the transformation preprocessing step and utilize the linearization, instead of the natural language description, of each modality as input. In addition, the summarization modality is also removed, since it is generated by LLM.
    \item \ours-gating: In \ours-gating, we use all modalities instead of just disease modality to generate the weights for the multimodal data integration component.
\end{itemize}
The results are shown in Table~\ref{tab:data_augment_ablation_study}, and the results of \ours\ are also reported for comparison. From those tables, we observe that: (1) \ours\ outperforms all the variants without certain components, including \ours-aug, \ours-aux and \ours-LLM, showcasing the effectiveness and complementary of the representation augmentation component, the auxiliary loss component and the LLM transformation preprocessing step; (2) \ours\ outperforms its variant, \ours-gating, with a slight advantage in performance. This suggests that determining the modality importance for each trial based solely on disease information is sufficient. Including additional modality information, even to a slight extent, appears to have a negative impact on performance.

\subsection{Analysis of Multimodal Data Integration}
\label{subsubsec:mix_of_experts}

We further analyze how multimodal data integration contributes to clinical outcome prediction. Here, we compare the performance of models using data from only one modality with \ours\ that integrates all those modalities. We report the results in Table~\ref{tab:moe_method_ablation_study}. The results indicate that \ours\ outperforms almost all unimodal models, demonstrating the effectiveness of multimodal integration.
In addition, the results also demonstrate that the drug description and the criteria modalities are the least and the most important modality, respectively, which is expected since the quality of recruited patients plays a crucial role in trial success~\cite{jin2017re,zhang2021ddn2}.

\subsection{Analysis of Sparse Mixture-of-Experts}

\begin{figure}
    \centering
    \includegraphics[width=0.9\columnwidth]{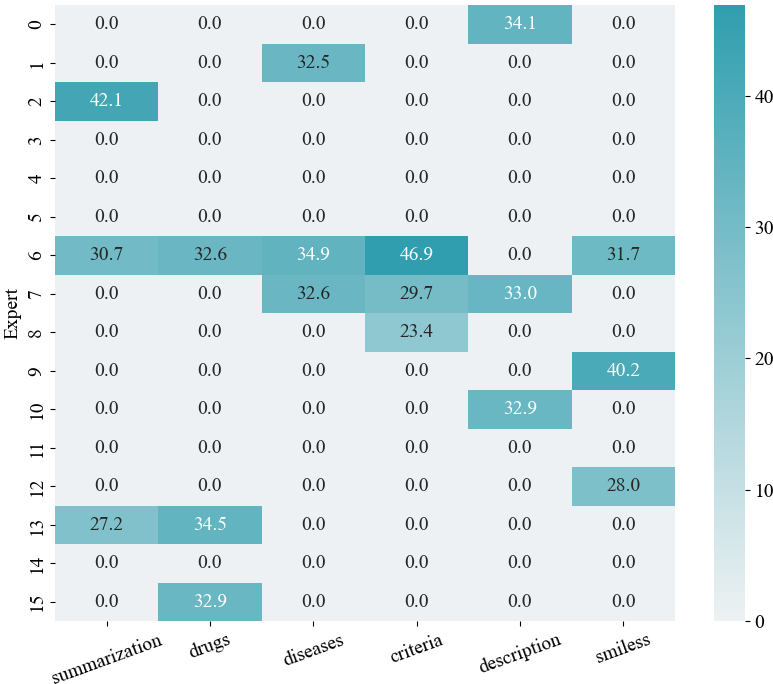}
    \caption{The SMoE experts' importance weights of our model predicting the knee osteoarthritis patient. Experts 6 and 7 play a crucial role in extracting common information patterns across modalities, while other experts specialize in a single specific modality.}
    \label{fig:smoe_weights}
\end{figure}

In addition, we delve into an analysis of the Sparse MoE model to understand the performance enhancements obtained by it. Here, we select a knee osteoarthritis patient case. For each modality, the SMoE framework selects top-$3$ experts from a pool of $16$ experts with the highest weights. The weights of these selected SMoE experts are visualized in Figure~\ref{fig:smoe_weights}. As expected, certain experts, such as 6 and 7, are consistently chosen across multiple modalities, indicating their pivotal role in extracting similar information patterns among different modalities. Furthermore, other experts demonstrate a more focused expertise, concentrating on one or two modalities. This demonstrates the effectiveness of the SMoE framework in both extracting similar information patterns across different modalities and capturing specialized information patterns within a single modality.

\subsection{Case Study}

\begin{figure}
    \centering
    \includegraphics[width=0.7\columnwidth]{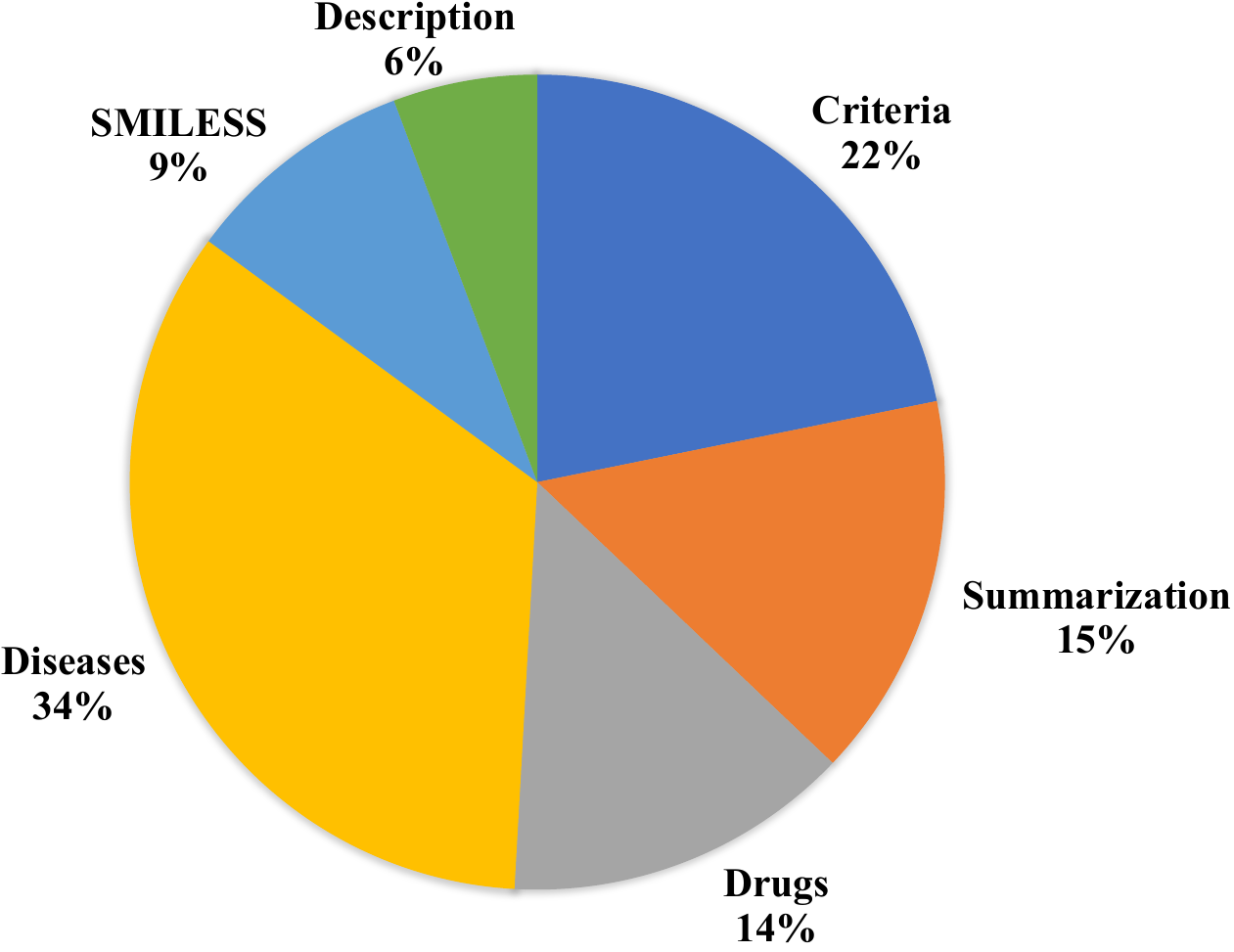}
    \caption{The modality importance weights of our model predicting the type 2 diabetes mellitus patient. \ours\ pay more attention to the disease modality as expected, since type 2 diabetes mellitus is hard to cure.}
    \label{fig:case_study}
    \vspace{-1em}
\end{figure}

In addition, we conduct a case study to analyze the contribution of each modality in clinical trial outcome prediction. Specifically, we analyze the result of a type 2 diabetes mellitus patient, who was inadequately controlled with metformin at the maximal effective and tolerated dose of metformin for at least 12 weeks. Since type 2 diabetes mellitus is hard to cure~\cite{chang2019integrated}, the model should pay attention to the name of the disease and predict the trial as failed, which is consistent with the behavior of our model. The modality importance weights are shown in Figure~\ref{fig:case_study}. As we expected, the attention weights of the disease modality are much higher than other modalities, which demonstrates that our model pays attention to the disease modality and predicts the trial correctly.

\section{Related Works}
\label{sec:Related_Works}

\textbf{Clinical Trial Outcome Prediction.}
Machine learning methods have been proven efficient on diverse tabular data prediction tasks, especially the clinical trial outcome prediction task, resulting in profound performances~\cite{chen2023excelformer,yan2023t2g}.
Recently, Fu \textit{et al.}~\shortcite{fu2023automated} proposed a hierarchical interaction network employing different encoders to fuse multiple modal data and capture their correlations for trial outcome predictions;
Wang \textit{et al.}~\shortcite{wang2023spot} clustered multi-sourced trial data into different topics, organizing trial embeddings for prediction. Wang \textit{et al.}~\shortcite{wang2023anypredict} converted clinical trial data into a format compatible description for prediction. However, converting all modalities into a single description poses significant challenges. This approach makes it difficult for the model to distinguish the unique information of each modality and necessitates external data to aid in differentiating these modalities. In contrast, \ours\ extracts representations for each modality separately and dynamically integrates them, providing a more effective way to preserve distinct characteristics of each modality.

\noindent \textbf{Mixture-of-Experts.}
Mixture-of-Experts (MoE) is a special type of neural network whose parameters are partitioned into a series of sub-modules, called experts, functioning in a conditional computation fashion~\cite{jacobs1991adaptive,jordan1993hierarchical}.
Recently, Shazzer \textit{et al.}~\shortcite{shazeer2017outrageously} simplified the MoE layer by selecting a sparse combination of the experts, instead of all experts, to process input data, significantly reducing the computational cost and improving the training stability.
To encourage specialization and decrease redundancy among experts~\cite{chen2022task}, Dai \textit{et al.}~\shortcite{dai2022one} pre-defined the expert assignment for different input categories, and Hazimeh \textit{et al.}~\shortcite{hazimeh2021dselect} advocated multiple, diverse router policies, facilitating the intriguing goals of SMoE is to divide and conquer the learning task by solving each piece of the task with adaptively selected experts~\cite{aoki2022heterogeneous,mittal2022modular}.
To identify similar information patterns between different modalities and extract them with the same expert model, \ours\ follows the design of Sparse Mixture-of-Experts~\cite{shazeer2017outrageously}, routing inputs to a subset of experts, dynamically selecting the experts instead of using a pre-defined assignment.

\section{Conclusion}
\label{sec:Conclusion}
We introduce \ours, an approach that unifies multimodal data using natural language descriptions and integrates this information within a Mixture-of-Experts (MoE) framework for clinical trial outcome prediction. We employ noise-resilient encoders to extract representations from each modality, utilize a Sparse MoE framework to further dig the similar information patterns in different modalities, and
introduce an auxiliary loss to improve the quality of modal-specific representations. Empirically, our LIFTED method demonstrates superior performance compared to existing approaches across all three phases of clinical trials, underscoring the effectiveness and potent representational capacity of natural language and highlighting the potential for a unified text modality to supplant diverse information modalities.

\bibliographystyle{named}
\bibliography{ijcai25}

\clearpage

\appendix

\renewcommand{\thefigure}{\thesection.\arabic{figure}}
\renewcommand{\thetable}{\thesection.\arabic{table}}
\setcounter{figure}{0}
\setcounter{table}{0}

\setcounter{linenumber}{0}
\renewcommand{\thelinenumber}{\arabic{linenumber}}

\section{Prompt}
\label{sec:prompt}

The whole prompt, including the system message, is demonstrated in Table~\ref{tab:prompting}, and some examples are demonstrated in Table~\ref{tab:examples_of_prompting}.

\begin{table}[htbp]
    \caption{Prompting.}
    \label{tab:prompting}
    \begin{tabular}{p{0.9\columnwidth}}
        \toprule
        \textbf{System Message}                                                             \\
        You are a helpful assistant.                                                        \\
        \midrule
        \textbf{Prompting}                                                                  \\
        Here is the schema definition of the table:

        \${schema\_definition}

        This is a sample from the table:

        \${linearization}

        Please briefly summarize the sample with its value in one sentence. You should describe the important values, like drugs and diseases, instead of just the names of columns in the table.

        A brief summarization of another sample may look like:

        This study will test the ability of extended-release nifedipine (Procardia XL), a blood pressure medication, to permit a decrease in the dose of glucocorticoid medication children take to treat congenital adrenal hyperplasia (CAH).

        Note that the example is not the summarization of the sample you have to summarize. \\
        \midrule
        \textbf{Response}                                                                   \\
        \${summarization\_of\_the\_sample}                                                  \\
        \bottomrule
    \end{tabular}
\end{table}

\begin{table*}[htbp]
    \caption{Examples of Prompting.}
    \label{tab:examples_of_prompting}
    \begin{tabular}{p{0.5\linewidth}p{0.4\linewidth}}
        \toprule
        Linearization                                                                                                                                                                                                                                                                                                                                                                                                                                                                                                                                                                                                                                                                                                                                                                                                                                                                                                                                                                      & Summarization                                                                                                                                                                                                                                                                                                                                                                                                                                                                                                                                                \\
        \midrule
        phase: phase 1/phase 2; diseases: ['adenocarcinoma of the lung', 'non-small cell lung cancer']; icdcodes: ["['D02.20', 'D02.21', 'D02.22']", "['C78.00', 'C78.01', 'C78.02', 'D14.30', 'D14.31', 'D14.32', 'C34.2']"]; drugs: ['erlotinib hydrochloride', 'hsp90 inhibitor auy922']; criteria: \textbackslash n        Inclusion Criteria:\textbackslash n          -  All patients must have pathologic evidence of advanced lung adenocarcinoma (stage IIIBor stage IV) confirmed histologically/cytologically at NU, MSKCC, or DFCI and EITHER previous RECIST-defined response $\dots$                                                                                                                                                                                                                                                                                                                                                                                         & This sample is a phase 1/phase 2 trial for patients with advanced lung adenocarcinoma, testing the efficacy of erlotinib hydrochloride and hsp90 inhibitor auy922 in patients who have previously responded to erlotinib or gefitinib or have a documented mutation in the EGFR gene. The study has specific inclusion and exclusion criteria, and patients must meet certain medical conditions and have negative pregnancy tests to be eligible.                                                                                                           \\
        \midrule
        phase: phase 2; diseases: ['multiple myeloma']; icdcodes: ["['C90.01', 'C90.02', 'C90.00']"]; drugs: ['dexamethasone', 'thalidomide', 'lenalidomide']; criteria: \textbackslash n        Inclusion Criteria:\textbackslash n\textbackslash n          -  Subject must voluntarily sign and understand written informed consent.\textbackslash n\textbackslash n          -  Age $>$ 18 years at the time of signing the consent form.\textbackslash n\textbackslash n          -  Histologically confirmed Salmon-Durie stage II or III MM. Stage I MM patients will be\textbackslash n             eligible if they display poor prognostic factors (ß2M $\geq$ 5.5 mg/L, plasma cell\textbackslash n             proliferation index $\geq$ 5\%, albumin of less than 3.0, and unfavorable cytogenetics). $\dots$                                                                                                                                                                & This sample is a phase 2 clinical trial for patients with relapsed or refractory multiple myeloma, testing the combination of dexamethasone, thalidomide, and lenalidomide as a treatment option. The eligibility criteria include specific disease stage, prior treatment history, and certain laboratory parameters. Exclusion criteria include non-secretory MM, prior history of other malignancies, and certain medical conditions.                                                                                                                     \\
        \midrule
        phase: phase 3; diseases: ["Alzheimer's disease"]; icdcodes: ["['G30.8', 'G30.9', 'G30.0', 'G30.1']"]; drugs: ['rivastigmine 5 cm\^{}2 transdermal patch', 'rivastigmine 10 cm\^{}2 transdermal patch']; criteria: \textbackslash n        Inclusion Criteria:\textbackslash n\textbackslash n          -  Be at least 50 years of age;\textbackslash n\textbackslash n          -  Have a diagnosis of probable Alzheimer's Disease; \textbackslash n\textbackslash n          -  Have an MMSE score of $\geq$ 10 and $\leq$ 24;\textbackslash n\textbackslash n          -  Must have a caregiver who is able to attend all study visits;\textbackslash n\textbackslash n          -  Have received continuous treatment with donepezil for at least 6 months prior to\textbackslash n             screening, and received a stable dose of 5 mg/day or 10 mg/day for at least the last 3\textbackslash n             of these 6 months.\textbackslash n\textbackslash n $\dots$ & This sample is a phase 3 clinical trial for Alzheimer's disease, testing the efficacy of rivastigmine transdermal patches in patients aged 50 and above with a diagnosis of probable Alzheimer's disease and an MMSE score between 10 and 24. The inclusion criteria also require patients to have a caregiver who can attend all study visits and have received continuous treatment with donepezil for at least 6 months prior to screening. The exclusion criteria include various medical conditions and disabilities that may interfere with the study. \\
        \bottomrule
    \end{tabular}
\end{table*}

\section{Baselines}
\label{sec:baselines}

Many methods have been selected as baselines in our experiments, including both statistical machine learning and deep learning models. We use the same setups in Fu \textit{et al.}~\shortcite{fu2022hint} and Wang \textit{et al.}~\shortcite{wang2023spot} for most of them.

\begin{itemize}[leftmargin=*]
    \item \textbf{Logistic regression (LR)}~\cite{lo2019learning,kien2021predicting}: logistic regression with the default hyperparameters implemented by scikit-learn~\cite{pedregosa2011scikit}.
    \item \textbf{Random Forest (RF)}~\cite{lo2019learning,kien2021predicting}: similar to logistic regression, the random forest is also implemented by scikit-learn with the default hyperparameters~\cite{pedregosa2011scikit}.
    \item \textbf{XGBoost}~\cite{rajpurkar2020evaluation,kien2021predicting}: An implementation of gradient-boosted decision trees optimized for speed and performance.
    \item \textbf{Adaptive boosting (AdaBoost)}~\cite{fan20201platelet}: an adaptive boosting-based decision tree method implemented by scikit-learn~\cite{pedregosa2011scikit}.
    \item \textbf{k Nearest Neighbor (kNN) + RF}~\cite{lo2019learning}: a combined model using kNN to imputate missing data and predicting by random forests.
    \item \textbf{Feedforward Neural Network (FFNN)}~\cite{tranchevent2019deep}: a feedforward neural network that uses the same feature as HINT~\cite{fu2023automated}. The FFNN contains three fully-connected layers with hidden dimensions of 500 and 100, as well as a rectified linear unit (ReLU) activation layer to provide nonlinearity.
    \item \textbf{Multi-Modal Fusion (MMF)}: This technique amalgamates multi-modal data to arrive at a final prediction, employing both early fusion and late fusion strategies. In the early fusion approach, various modal inputs are first concatenated before being fed into the prediction model. Conversely, in the late fusion variant, multiple prediction models are employed on each modal input, and the ultimate prediction is derived through fusion techniques, such as voting, which integrates predictions from each modality.
    \item \textbf{HINT}~\cite{fu2022hint}: several key components are integrated with HINT, including a drug molecule encoder utilizing MPNN algorithm, a disease ontology encoder based on GRAM, a trial eligibility criteria encoder leveraging BERT, and also, a drug molecule pharmacokinetic encoder, surplus a graph neural network to capture feature interactions. After the interacted features are encoded, they are fed into a prediction model for accurate outcome predictions.
    \item \textbf{SPOT}~\cite{wang2023spot}: SPOT contains several steps. Firstly, trial topics are identified to group the diverse trial data from multiple sources into relevant trial topics. Subsequently, trial embeddings are produced and organized according to topic and timestamp to construct organized clinical trial sequences. Finally, each trial sequence is treated as a separate task, and a meta-learning approach is employed to adapt to new tasks with minimal modifications.
\end{itemize}

\section{Dataset Descriptions}
\label{sec:dataset_descriptions}

\begin{table}[h]
    \small
    \centering
    \caption{The statistics of the HINT Dataset and the CTOD Dataset. The number of Trials is shown by the split of train/validation/test sets.}
    \vspace{-0.5em}
    \label{tab:hint_dataset}
    \resizebox{\linewidth}{!}{
        \begin{tabular}{cccccc}
            \toprule
            \multicolumn{6}{c}{\textbf{HINT}}                                              \\
            \midrule
                      & \# Trials       & \# Drugs & \# Diseases & \# Success & \# Failure \\
            \midrule
            Phase I   & 1,044/116/627   & 2,020    & 1,392       & 1,006      & 781        \\
            Phase II  & 4,004/445/1,653 & 5,610    & 2,824       & 3,039      & 3,063      \\
            Phase III & 3,092/344/1,140 & 4,727    & 1,619       & 3,104      & 1,472      \\
            \midrule
            \multicolumn{6}{c}{\textbf{CTOD}}                                              \\
            \midrule
                      & \# Trials       & \# Drugs & \# Diseases & \# Success & \# Failure \\
            \midrule
            Phase I   & 1,012/149/627   & 1,638    & 913         & 1,129      & 659        \\
            Phase II  & 3,950/501/1,653 & 5,003    & 2,254       & 3,949      & 2,156      \\
            Phase III & 3,075/363/1,140 & 3,863    & 1,533       & 3,643      & 941        \\
            \bottomrule
        \end{tabular}}
\end{table}

The HINT dataset~\cite{fu2022hint,chen2024uncertainty} and CTOD dataset~\cite{gao2024automaticallylabeling200blifesaving} include the information on diseases, the name, description, and SMILES string of drugs, eligibility criteria for each clinical trial record, the phase, and also, the trial outcome labels as success or failure covering Phases I, II and III trials. The HINT dataset contains 17,538 clinical trial records, with 1,787 trials in Phase I, 6,102 trials in Phase II, and 4,576 trials in Phase III~\cite{fu2021probabilistic}. We utilize parts of the CTOD dataset to test \ours's performance, with 1,788 trials in Phase I, 6,105 trials in Phase II, and 4,584 trials in Phase III, for a total of 12,477 trials.~\cite{gao2024automaticallylabeling200blifesaving}. The detailed data statistics of the HINT dataset and the CTOD dataset are shown in Table~\ref{tab:hint_dataset}.

In our implementation, we incorporate all modalities, including disease, the name, description and SMILES string of drugs and criteria, totaling five modalities. Additionally, we include phase information when generating the natural language summarization for samples. The transform-based encoder for the SMILES string modality is pre-trained and the corresponding tokenizer is specifically designed for SMILES string data. However, all the other modalities are tokenized by a unified tokenizer and none of the other encoders are pre-trained.

\section{Hyperparameter Settings}

We follow the settings of most hyperparameters in HINT~\cite{fu2022hint}. The models are trained for a total of 5 epochs using a mini-batch size of 32 on one NVIDIA 4090 GPUs, which will take up to 2 hours. We employ the AdamW optimizer~\cite{Loshchilov2017DecoupledWD} with a learning rate of $3 \times 10^{-4}$, $\beta$ values of $(0.9, 0.99)$, and a weight decay of $1 \times 10^{-2}$ with a CosineAnnealing learning rate scheduler.

\section{Detailed Results}
\label{sec:detailed_results}

\begin{table*}[h]
    \caption{The clinical trial outcome performance (\%) of \ours\ and baselines on HINT dataset and CTOD dataset. The mean and standard deviations are calculated from 30 independent runs with different random seeds. $^\dagger$: The results of the HINT and SPOT methods were obtained by running their released codes. The best results and second best results are \textbf{bold} and \underline{underlined}, respectively. We observe that \ours\ consistently outperforms all other methods over all three phases.}
    \vspace{-0.5em}
    \label{tab:performance_phase_detail}
    \resizebox{\linewidth}{!}{
        \begin{tabular}{l|c|c|c|c|c|c|c|c|c}
            \toprule
            \multicolumn{10}{c}{\textbf{HINT}}                                                                                                                                                                                                                                                                                                                         \\
            \midrule
                                       & \multicolumn{3}{c|}{\textbf{Phase I Trials}} & \multicolumn{3}{c|}{\textbf{Phase II Trials}} & \multicolumn{3}{c}{\textbf{Phase III Trials}}                                                                                                                                                                                  \\
            \cmidrule{2-10}
            \textbf{Method}            & \textbf{PR-AUC}                              & \textbf{F1}                                   & \textbf{ROC-AUC}                              & \textbf{PR-AUC}            & \textbf{F1}                 & \textbf{ROC-AUC}           & \textbf{PR-AUC}              & \textbf{F1}                & \textbf{ROC-AUC}           \\
            \midrule
            LR                         & $50.0 \pm 0.5$                               & $60.4 \pm 0.5$                                & $52.0 \pm 0.6$                                & $56.5 \pm 0.5$             & $55.5 \pm 0.6$              & $58.7 \pm 0.9$             & $68.7 \pm 0.5$               & $69.8 \pm 0.5$             & $65.0 \pm 0.7$             \\
            RF                         & $51.8 \pm 0.5$                               & $62.1 \pm 0.5$                                & $52.5 \pm 0.6$                                & $57.8 \pm 0.8$             & $56.3 \pm 0.9$              & $58.8 \pm 0.9$             & $69.2 \pm 0.4$               & $68.6 \pm 1.0$             & $66.3 \pm 0.7$             \\
            XGBoost                    & $51.3 \pm 6.0$                               & $62.1 \pm 0.7$                                & $51.8 \pm 0.6$                                & $58.6 \pm 0.6$             & $57.0 \pm 0.9$              & $60.0 \pm 0.7$             & $69.7 \pm 0.7$               & $69.6 \pm 0.5$             & $66.7 \pm 0.5$             \\
            AdaBoost                   & $51.9 \pm 0.5$                               & $62.2 \pm 0.7$                                & $52.6 \pm 0.6$                                & $58.6 \pm 0.9$             & $58.3 \pm 0.8$              & $60.3 \pm 0.7$             & $70.1 \pm 0.5$               & $69.5 \pm 0.5$             & $67.0 \pm 0.4$             \\
            kNN+RF                     & $53.1 \pm 0.6$                               & $62.5 \pm 0.7$                                & $53.8 \pm 0.5$                                & $59.4 \pm 0.8$             & $59.0 \pm 0.6$              & $59.7 \pm 0.8$             & $70.7 \pm 0.7$               & $69.8 \pm 0.8$             & $67.8 \pm 1.0$             \\\midrule
            FFNN                       & $54.7 \pm 1.0$                               & $63.4 \pm 1.5$                                & $55.0 \pm 1.0$                                & $60.4 \pm 1.0$             & $59.9 \pm 1.2$              & $61.1 \pm 1.1$             & $74.7 \pm 1.1$               & $74.8 \pm 0.9$             & $68.1 \pm 0.8$             \\
            MMF (early fusion)         & $60.6 \pm 2.8$                               & $59.4 \pm 2.3$                                & $54.4 \pm 2.4$                                & $60.2 \pm 1.9$             & $62.6 \pm 1.4$              & $60.7 \pm 1.3$             & $85.5 \pm 1.4$               & $81.5 \pm 0.9$             & $70.6 \pm 1.7$             \\
            MMF (late fusion)          & $63.4 \pm 3.0$                               & $67.5 \pm 2.2$                                & $59.0 \pm 2.8$                                & \underline{$62.9 \pm 2.0$} & $63.0 \pm 1.5$              & $62.6 \pm 1.6$             & \underline{$86.9 \pm 1.6$}   & \underline{$83.1 \pm 1.1$} & \underline{$71.8 \pm 2.2$} \\
            DeepEnroll                 & $56.8 \pm 0.7$                               & $64.8 \pm 1.1$                                & $57.5 \pm 1.3$                                & $60.0 \pm 1.0$             & $59.8 \pm 0.7$              & $62.5 \pm 0.8$             & $77.7 \pm 0.8$               & $78.6 \pm 0.7$             & $69.9 \pm 0.8$             \\
            COMPOSE                    & $56.4 \pm 0.7$                               & $65.8 \pm 0.9$                                & $57.1 \pm 1.1$                                & $60.4 \pm 0.7$             & $59.7 \pm 0.6$              & $62.8 \pm 0.9$             & $78.2 \pm 0.8$               & $79.2 \pm 0.7$             & $70.0 \pm 0.7$             \\
            HINT$^\dagger$             & $58.4 \pm 2.3$                               & $68.2 \pm 1.7$                                & $62.1 \pm 2.2$                                & $59.1 \pm 1.2$             & $63.9 \pm 1.2$              & $62.8 \pm 1.4$             & \underline{$85.9 \pm 1.1$}   & $80.9 \pm 0.8$             & $70.8 \pm 1.3$             \\
            SPOT$^\dagger$             & \underline{$69.8 \pm 1.7$}                   & \underline{$68.4 \pm 1.2$}                    & \underline{$64.6 \pm 2.1$}                    & $62.6 \pm 0.7$             & \underline{$64.3 \pm 0.6$}  & \underline{$63.0 \pm 0.6$} & $81.7 \pm 0.8$               & $81.0 \pm 0.4$             & $71.0 \pm 0.4$             \\
            \midrule
            \textbf{\ours\ (ours)}     & $\textbf{70.7 $\pm$ 2.3}$                    & $\textbf{71.6 $\pm$ 1.4}$                     & $\textbf{64.9 $\pm$ 2.1}$                     & $\textbf{69.8 $\pm$ 1.8}$  & $\textbf{66.2 $\pm$ 1.1}$   & $\textbf{65.1 $\pm$ 1.4}$  & $\textbf{88.3 $\pm$ 1.1}$    & $\textbf{83.8 $\pm$ 0.8}$  & $\textbf{73.5 $\pm$ 1.6}$  \\
            \midrule
            \multicolumn{10}{c}{\textbf{CTOD}}                                                                                                                                                                                                                                                                                                                         \\
            \midrule
                                       & \multicolumn{3}{c|}{\textbf{Phase I Trials}} & \multicolumn{3}{c|}{\textbf{Phase II Trials}} & \multicolumn{3}{c}{\textbf{Phase III Trials}}                                                                                                                                                                                  \\
            \cmidrule{2-10}
            \textbf{Method}            & \textbf{PR-AUC}                              & \textbf{F1}                                   & \textbf{ROC-AUC}                              & \textbf{PR-AUC}            & \textbf{F1}                 & \textbf{ROC-AUC}           & \textbf{PR-AUC}              & \textbf{F1}                & \textbf{ROC-AUC}           \\
            \midrule
            LR                         &
            $55.6 \pm 2.3$             & $58.5 \pm 2.1$                               & $51.5 \pm 2.2$                                &
            $56.0 \pm 1.4$             & $63.6 \pm 1.1$                               & $56.7 \pm 1.0$                                &
            $75.0 \pm 2.2$             & $73.8 \pm 1.6$                               & $51.9 \pm 2.0$                                                                                                                                                                                                                                                                 \\
            RF                         &
            $58.3 \pm 2.9$             & $70.0 \pm 1.6$                               & $54.0 \pm 2.0$                                &
            $60.8 \pm 1.3$             & $70.9 \pm 0.8$                               & $56.0 \pm 1.3$                                &
            $75.1 \pm 1.6$             & $85.1 \pm 0.7$                               & $50.7 \pm 2.2$
            \\
            XGBoost                    &
            $57.4 \pm 3.5$             & $68.0 \pm 1.9$                               & $50.5 \pm 2.3$                                &
            $61.2 \pm 1.5$             & $70.6 \pm 1.2$                               & $61.3 \pm 1.3$                                &
            $78.7 \pm 1.8$             & $84.9 \pm 0.7$                               & $57.1 \pm 2.0$                                                                                                                                                                                                                                                                 \\
            AdaBoost                   &
            $58.8 \pm 3.2$             & $66.7 \pm 2.4$                               & $53.5 \pm 2.3$                                &
            $60.2 \pm 1.8$             & $68.9 \pm 0.8$                               & $55.2 \pm 1.3$                                &
            $79.5 \pm 1.3$             & $84.6 \pm 0.8$                               & $56.9 \pm 1.9$                                                                                                                                                                                                                                                                 \\
            kNN+RF                     &
            $58.4 \pm 3.9$             & $70.3 \pm 2.0$                               & $58.7 \pm 2.1$                                &
            $61.2 \pm 1.7$             & $70.9 \pm 0.9$                               & $56.6 \pm 1.5$                                &
            $76.8 \pm 2.2$             & $85.2 \pm 1.1$                               & $52.5 \pm 1.9$                                                                                                                                                                                                                                                                 \\\midrule
            {FFNN}                     &
            $55.2 \pm 2.2$             & $58.4 \pm 1.4$                               & $48.7 \pm 1.7$                                &
            $57.2 \pm 1.5$             & $66.0 \pm 1.3$                               & $53.0 \pm 1.2$                                &
            $76.1 \pm 1.4$             & $79.6 \pm 1.1$                               & $52.6 \pm 1.5$
            \\
            {MMF (early fusion)}       &
            {$61.2 \pm 2.8 $ }         & {$64.2 \pm 2.1 $ }                           & {$56.2 \pm 2.8$}                              &
            $64.0 \pm 1.2$             & $70.1 \pm 0.9$                               & $59.0 \pm 1.1$                                &
            {$83.3 \pm 1.4$}           & {$84.4 \pm 1.0$}                             & {$64.6 \pm 1.9$   }                                                                                                                                                                                                                                                            \\
            {MMF (late fusion)}        &
            $63.0 \pm 3.1$             & $70.5 \pm 1.8$                               & $57.4 \pm 2.4$                                &
            \underline{$65.7 \pm 2.1$} & {$71.4 \pm 0.8$}                             & $60.3 \pm 2.6$                                &
            \underline{$83.5 \pm 1.7$} & {$85.2 \pm 0.8$}                             & {$65.8 \pm 1.6$}
            \\
            HINT$^\dagger$             &
            $63.4 \pm 2.9$             & \underline{$71.0 \pm 1.2$}                   & $57.6 \pm 2.4$                                & $64.6 \pm 2.4$                                & $71.2 \pm 1.3$             & \underline{$60.6 \pm 2.0$ } & $81.8 \pm 1.9$             & \underline{$85.7 \pm 0.7$  } & $60.8 \pm 2.0$                                          \\
            SPOT$^\dagger$             &
            \underline{$66.8 \pm 1.9$} & {$70.0 \pm 0.9$}                             & \underline{$62.3 \pm 1.0$}                    &

            {$64.5 \pm 1.7$}           & \underline{$71.8 \pm 0.7$}                   & {$58.8 \pm 1.4$}                              &

            $83.2 \pm 2.3$             & $77.6 \pm 1.0$                               & \underline{$67.5 \pm 1.7$ }
            \\
            \midrule
            {\textbf{\ours\ (ours)}}   &
            $\textbf{69.7 $\pm$ 3.0}$  & $\textbf{71.8 $\pm$ 1.4}$                    & $\textbf{63.4 $\pm$ 2.3}$                     &
            $\textbf{67.7 $\pm$ 2.0}$  & $\textbf{72.0 $\pm$ 1.4}$                    & $\textbf{63.0 $\pm$ 1.2}$                     &
            $\textbf{86.7 $\pm$ 1.2}$  & $\textbf{85.9 $\pm$ 1.0}$                    & $\textbf{69.5 $\pm$ 1.8}$                                                                                                                                                                                                                                                      \\
            \bottomrule
        \end{tabular}}
    \vspace{-0.5em}

\end{table*}

\begin{table*}[h]
    \caption{The clinical trial outcome prediction performance (\%) of \ours\ and variants without certain key component. The best results are \textbf{bold}. \ours\ outperforms all variants, showcasing the effectiveness of our proposed components.}
    \vspace{-0.5em}
    \label{tab:data_augment_ablation_study_detail}
    \resizebox{\linewidth}{!}{
        \begin{tabular}{l|c|c|c|c|c|c|c|c|c}
            \toprule
            \multicolumn{10}{c}{\textbf{HINT}}                                                                                                                                                                                                                                                                                                             \\
            \midrule
                                      & \multicolumn{3}{c|}{\textbf{Phase I Trials}} & \multicolumn{3}{c|}{\textbf{Phase II Trials}} & \multicolumn{3}{c}{\textbf{Phase III Trials}}                                                                                                                                                                       \\
            \cmidrule{2-10}
            \textbf{Method}           & \textbf{PR}                                  & \textbf{F1}                                   & \textbf{ROC}                                  & \textbf{PR}               & \textbf{F1}               & \textbf{ROC}              & \textbf{PR}               & \textbf{F1}               & \textbf{ROC}            \\
            \midrule

            \ours-aug
                                      & $68.4 \pm 2.0$                               & $69.8 \pm 2.1$                                & $64.8 \pm 1.5$
                                      & $69.5 \pm 1.4$                               & $66.0 \pm 1.1$                                & $64.3 \pm 0.8$
                                      & $86.9 \pm 1.7$                               & $82.4 \pm 0.9$                                & $72.1 \pm 1.6$                                                                                                                                                                                                      \\

            \ours-aux                 & $69.0 \pm 2.9$                               & $71.2 \pm 1.7$                                & $63.7 \pm 1.6$                                &
            $69.6 \pm 1.6$            & $64.5 \pm 1.5$                               & $64.6 \pm 1.3$                                &
            $87.4 \pm 1.4$            & $82.8 \pm 1.0$                               & $71.1 \pm 2.2$                                                                                                                                                                                                                                                      \\

            \ours-LLM                 & $68.5 \pm 2.7$                               & $70.8 \pm 1.3$                                & $64.0 \pm 2.3$                                & $69.7 \pm 2.0$            & $64.9 \pm 1.4$            & $65.0 \pm 1.5$            & $86.7 \pm 1.0$            & $82.7 \pm 1.0$            & $70.8 \pm 1.3$          \\

            \ours-gating              & 69.9 $\pm$ 2.3                               & 71.3 $\pm$ 1.8                                & 64.9 $\pm$ 1.9
                                      & 69.7 $\pm$ 1.7                               & 65.5 $\pm$ 1.4                                & 65.0 $\pm$ 1.6
                                      & 87.0 $\pm$ 0.8                               & 82.7 $\pm$ 0.8                                & 72.4 $\pm$ 1.1                                                                                                                                                                                                      \\
            \midrule

            \textbf{\ours\ (ours)}    & $\textbf{70.7 $\pm$ 2.3}$                    & $\textbf{71.6 $\pm$ 1.4}$                     & $\textbf{64.9 $\pm$ 2.1}$                     & $\textbf{69.8 $\pm$ 1.8}$ & $\textbf{66.2 $\pm$ 1.1}$ & $\textbf{65.1 $\pm$ 1.4}$ & $\textbf{88.3 $\pm$ 1.1}$ & $\textbf{83.8 $\pm$ 0.8}$ & \textbf{73.5 $\pm$ 1.6} \\
            \midrule
            \multicolumn{10}{c}{\textbf{CTOD}}                                                                                                                                                                                                                                                                                                             \\
            \midrule
                                      & \multicolumn{3}{c|}{\textbf{Phase I Trials}} & \multicolumn{3}{c|}{\textbf{Phase II Trials}} & \multicolumn{3}{c}{\textbf{Phase III Trials}}                                                                                                                                                                       \\
            \cmidrule{2-10}
            \textbf{Method}           & \textbf{PR}                                  & \textbf{F1}                                   & \textbf{ROC}                                  & \textbf{PR}               & \textbf{F1}               & \textbf{ROC}              & \textbf{PR}               & \textbf{F1}               & \textbf{ROC}            \\
            \midrule
            {\ours-aug}               &
            $65.1 \pm 3.7$            & $70.6 \pm 1.7$                               & $61.9 \pm 2.4$                                &
            $67.1 \pm 1.8$            & $70.5 \pm 1.1$                               & $62.2 \pm 1.4$                                &
            $85.2 \pm 1.4$            & $82.4 \pm 1.1$                               & $67.8 \pm 1.6$                                                                                                                                                                                                                                                      \\

            {\ours-aux}               &
            $64.4 \pm 3.3$            & $71.1 \pm 1.8$                               & $60.0 \pm 2.8$                                & $60.0 \pm 1.4$                                & $71.0 \pm 0.9$            & $54.9 \pm 1.0$            & $83.8 \pm 1.8$            & $85.5 \pm 0.9$            & $64.6 \pm 2.4$                                      \\

            {\ours-LLM}               &
            $65.1 \pm 2.7$            & $71.4 \pm 1.6$                               & $58.5 \pm 3.2$                                & $66.3 \pm 1.9$                                & $71.1 \pm 1.1$            & $61.2 \pm 1.5$            & $83.9 \pm 1.3$            & $85.4 \pm 1.0$            & $65.3 \pm 2.0$                                      \\

            {\ours-gating}            &
            67.1 $\pm$ 2.4            & 71.4 $\pm$ 1.8                               & 60.8 $\pm$ 2.5                                &
            65.1 $\pm$ 1.4            & 70.4 $\pm$ 1.3                               & 61.9 $\pm$ 1.5                                &
            85.0 $\pm$ 1.5            & 85.8 $\pm$ 0.7                               & 68.8 $\pm$ 1.8                                                                                                                                                                                                                                                      \\
            \midrule
            \textbf{\ours\ (ours)}    &
            $\textbf{69.7 $\pm$ 3.0}$ & $\textbf{71.8 $\pm$ 1.4}$                    & $\textbf{63.4 $\pm$ 2.3}$                     &
            $\textbf{67.7 $\pm$ 2.0}$ & $\textbf{72.0 $\pm$ 1.4}$                    & $\textbf{63.0 $\pm$ 1.2}$                     &
            $\textbf{86.7 $\pm$ 1.2}$ & $\textbf{85.9 $\pm$ 1.0}$                    & $\textbf{69.5 $\pm$ 1.8}$                                                                                                                                                                                                                                           \\
            \bottomrule
        \end{tabular}}
    \vspace{-0.5em}
\end{table*}

\begin{table*}[h]
    \caption{Performance analysis of multimodal data integration. The best results and second best results are \textbf{bold} and \underline{underlined}, respectively.}
    \vspace{-0.5em}
    \label{tab:moe_method_ablation_study_detail}
    \resizebox{\linewidth}{!}{
        \begin{tabular}{l|c|c|c|c|c|c|c|c|c}
            \toprule
            \multicolumn{10}{c}{\textbf{HINT}}                                                                                                                                                                                                                                                                                                                      \\
            \midrule
                                       & \multicolumn{3}{c|}{\textbf{Phase I Trials}} & \multicolumn{3}{c|}{\textbf{Phase II Trials}} & \multicolumn{3}{c}{\textbf{Phase III Trials}}                                                                                                                                                                               \\
            \cmidrule{2-10}
                                       & \textbf{PR-AUC}                              & \textbf{F1}                                   & \textbf{ROC-AUC}                              & \textbf{PR-AUC}            & \textbf{F1}                & \textbf{ROC-AUC}           & \textbf{PR-AUC}            & \textbf{F1}                & \textbf{ROC-AUC}           \\
            \midrule
            Summarization              & $63.2 \pm 2.4$                               & $69.9 \pm 2.2$                                & $57.8 \pm 2.2$                                & $66.1 \pm 1.3$             & $61.2 \pm 1.5$             & $61.2 \pm 1.0$             & $85.1 \pm 1.0$             & $80.7 \pm 1.0$             & $66.6 \pm 1.6$             \\
            Drugs                      & $62.1 \pm 1.2$                               & $67.2 \pm 1.5$                                & $57.9 \pm 1.1$                                & $60.5 \pm 1.1$             & $62.3 \pm 1.5$             & $55.8 \pm 1.1$             & $83.8 \pm 0.7$             & $81.7 \pm 1.0$             & $63.8 \pm 1.3$             \\
            Disease                    & $65.3 \pm 1.1$                               & $67.3 \pm 1.8$                                & $59.7 \pm 1.3$                                & \underline{$68.0 \pm 0.5$} & $59.9 \pm 1.3$             & $62.4 \pm 0.6$             & \underline{$86.0 \pm 0.8$} & $80.5 \pm 1.1$             & \underline{$69.1 \pm 0.9$} \\
            Description                & $55.5 \pm 0.5$                               & \underline{71.3 $\pm$ 0.1}                    & $50.2 \pm 1.0$                                & $55.5 \pm 0.7$             & $ 0.0 \pm 0.0$             & $50.0 \pm 1.3$             & $74.9 \pm 0.6$             & \textbf{85.7 $\pm$ 0.0}    & $49.7 \pm 1.3$             \\
            SMILES                     & $62.8 \pm 0.7$                               & $69.6 \pm 1.7$                                & $58.5 \pm 0.9$                                & $59.3 \pm 0.6$             & $58.3 \pm 2.2$             & $54.9 \pm 0.7$             & $76.1 \pm 1.5$             & $83.6 \pm 0.5$             & $51.1 \pm 2.5$             \\
            Criteria                   & \underline{$68.0 \pm 3.0$}                   & $70.5 \pm 1.9$                                & \underline{$63.1 \pm 2.1$}                    & $67.6 \pm 1.1$             & \underline{64.4 $\pm$ 1.0} & \underline{$63.0 \pm 1.3$} & $83.7 \pm 1.2$             & $82.7 \pm 0.7$             & $65.0 \pm 2.1$             \\
            \midrule
            \textbf{All (\ours)}       & $\textbf{70.7 $\pm$ 2.3}$                    & $\textbf{71.6 $\pm$ 1.4}$                     & $\textbf{64.9 $\pm$ 2.1}$                     & $\textbf{69.8 $\pm$ 1.8}$  & $\textbf{66.2 $\pm$ 1.1}$  & $\textbf{65.1 $\pm$ 1.4}$  & $\textbf{88.3 $\pm$ 1.1}$  & \underline{$83.8 \pm 0.8$} & $\textbf{73.5 $\pm$ 1.6}$  \\
            \midrule
            \multicolumn{10}{c}{\textbf{CTOD}}                                                                                                                                                                                                                                                                                                                      \\
            \midrule
                                       & \multicolumn{3}{c|}{\textbf{Phase I Trials}} & \multicolumn{3}{c|}{\textbf{Phase II Trials}} & \multicolumn{3}{c}{\textbf{Phase III Trials}}                                                                                                                                                                               \\
            \cmidrule{2-10}            & \textbf{PR-AUC}                              & \textbf{F1}                                   & \textbf{ROC-AUC}                              & \textbf{PR-AUC}            & \textbf{F1}                & \textbf{ROC-AUC}           & \textbf{PR-AUC}            & \textbf{F1}                & \textbf{ROC-AUC}           \\
            \midrule
            Summarization              &
            $61.6 \pm 3.2$             & $70.9 \pm 2.2$                               & $56.9 \pm 2.6$                                & {$65.4 \pm 1.3$}                              & \underline{$71.5 \pm 0.9$} & $61.1 \pm 1.3$             &
            $84.1 \pm 1.3$             & $84.8 \pm 0.9$                               & $64.9 \pm 1.9$                                                                                                                                                                                                                                                              \\
            Drugs                      &
            $60.8 \pm 2.8$             & $70.6 \pm 1.4$                               & $57.6 \pm 2.4$                                &
            $58.0 \pm 1.8$             & $71.3 \pm 1.1$                               & $53.8 \pm 1.2$                                & $77.3 \pm 1.7$                                & $85.2 \pm 1.0$             & $54.5 \pm 1.9$                                                                                                                                 \\
            Disease                    &
            \underline{$65.2 \pm 2.0$} & $70.2 \pm 1.9$                               & \underline{$61.3 \pm 1.7$}                    &
            \underline{$68.3 \pm 1.5$} & $71.1 \pm 1.2$                               & \underline{$62.6 \pm 1.0$}                    &
            \underline{$85.3 \pm 1.4$} & $85.5 \pm 0.7$                               & \underline{$68.4 \pm 2.0$}
            \\
            Description                &
            $56.6 \pm 2.6$             & $70.7 \pm 1.5$                               & $52.3 \pm 2.7$                                &
            $56.0 \pm 1.8$             & $ 71.3 \pm 0.8$                              & $52.4 \pm 1.7$                                &
            $76.5 \pm 1.2$             & \underline{85.7 $\pm$ 0.7}                   & $52.9 \pm 1.9 $                                                                                                                                                                                                                                                             \\
            SMILES                     &
            $61.6 \pm 2.8$             & $70.6 \pm 1.7$                               & $56.0 \pm 2.4$                                &
            $58.0 \pm 1.8$             & \underline{$71.5 \pm 1.2$}                   & $53.1 \pm 1.2$                                &
            $75.1 \pm 1.6$             & $85.6 \pm 0.7$                               & $50.8 \pm 2.4$                                                                                                                                                                                                                                                              \\
            Criteria                   &
            {$60.5 \pm 3.0$}           & \underline{$71.2 \pm 1.7$}                   & {$55.8 \pm 3.1$}                              &
            $59.9 \pm 2.5$             & {71.2$\pm$ 1.2}                              & {$53.9 \pm 2.0$}                              &
            $75.7 \pm 2.0$             & $84.6 \pm 0.7$                               & $52.5 \pm 2.6$                                                                                                                                                                                                                                                              \\
            \midrule
            \textbf{\ours\ (ours)}     &
            $\textbf{69.7 $\pm$ 3.0}$  & $\textbf{71.8 $\pm$ 1.4}$                    & $\textbf{63.4 $\pm$ 2.3}$                     &
            $\textbf{67.7 $\pm$ 2.0}$  & $\textbf{72.0 $\pm$ 1.4}$                    & $\textbf{63.0 $\pm$ 1.2}$                     &
            $\textbf{86.7 $\pm$ 1.2}$  & $\textbf{85.9 $\pm$ 1.0}$                    & $\textbf{69.5 $\pm$ 1.8}$                                                                                                                                                                                                                                                   \\
            \bottomrule
        \end{tabular}}
    \vspace{-0.5em}
\end{table*}

The detailed experiments results with variants are presented in Table~\ref{tab:performance_phase_detail}, \ref{tab:data_augment_ablation_study_detail}, \ref{tab:moe_method_ablation_study_detail}

\end{document}